\documentclass{bmvc2k}

\usepackage{times}
\usepackage{epsfig}
\usepackage{graphicx}
\usepackage{amsmath}
\usepackage{amssymb}
\usepackage{bm}
\usepackage{booktabs,colortbl,tabularx}
\usepackage{footmisc}

\usepackage{mathtools}
\usepackage{commath}

\usepackage{algorithm}
\usepackage[noend]{algpseudocode}
\usepackage{bbm}
\usepackage{comment}
\usepackage{multirow}
\usepackage{enumitem}

\usepackage{pifont}

\def\thefootnote{$\dagger$}\footnotetext{Corresponding author.}

\title{Rethinking Prototypical Contrastive Learning \\ through \\ Alignment, Uniformity and Correlation}

\addauthor{Shentong Mo}{shentonm@andrew.cmu.edu}{1}
\addauthor{Zhun Sun\thefootnote{}}{zhunsun@gmail.com}{2}
\addauthor{Chao Li}{chao.li@riken.jp}{3}
\addinstitution{
 Carnegie Mellon University\\
 Pittsburgh, PA 15213, United States
}
\addinstitution{
 Tohoku University\\
 Sendai, Miyagi, Japan
}
\addinstitution{
 Center for Advanced Intelligence Project (AIP), RIKEN \\
 Tokyo, Japan
}

\runninghead{Shentong Mo, Zhun Sun, and Chao Li}{PAUC}

\begin{document}

\maketitle

\begin{abstract}

Contrastive self-supervised learning (CSL) with a prototypical regularization has been introduced in learning meaningful representations for downstream tasks that require strong semantic information.
However, optimizing CSL with a loss that performs the prototypical regularization aggressively, \textit{e.g.}, the ProtoNCE loss, might cause the ``coagulation'' of examples in the embedding space. That is, the intra-prototype diversity of samples collapses to trivial solutions for their prototype being well-separated from others.
Motivated by previous works, we propose to mitigate this phenomenon by learning \textbf{P}rototypical representation through \textbf{A}lignment, \textbf{U}niformity and \textbf{C}orrelation \textbf{(\texttt{PAUC})}. Specifically, the ordinary ProtoNCE loss is revised with: (1) an alignment loss that pulls embeddings from positive prototypes together; (2) a uniformity loss that distributes the prototypical level features uniformly; (3) a correlation loss that increases the diversity and discriminability between prototypical level features.
We conduct extensive experiments on various benchmarks where the results demonstrate the effectiveness of our method on improving the quality of prototypical contrastive representations.
Particularly, in the classification down-stream tasks with linear probes, our proposed method outperforms the state-of-the-art instance-wise and prototypical contrastive learning methods on the ImageNet-100 dataset by 2.96\% and the ImageNet-1K dataset by 2.46\% under the same settings of batch size and epochs.
  
\end{abstract}

\section{Introduction}
\vspace{-0.5em}
Contrastive learning in a prototypical style~\cite{li2021prototypical,caron2020unsupervised,wang2021cld,mo2021spcl} has achieved remarkable progress in terms of learning meaningful representations under the self-supervised setting.
The major goal of introducing prototypes is to encourage representations to be closer to clusters of samples that contain certain semantic meaning (\textit{i.e.} positive prototypes), while being away from the clusters that do not \mbox{(\textit{i.e.} negative prototypes)}.

\begin{figure}[!htb]
\centering
\includegraphics[width=0.6\linewidth]{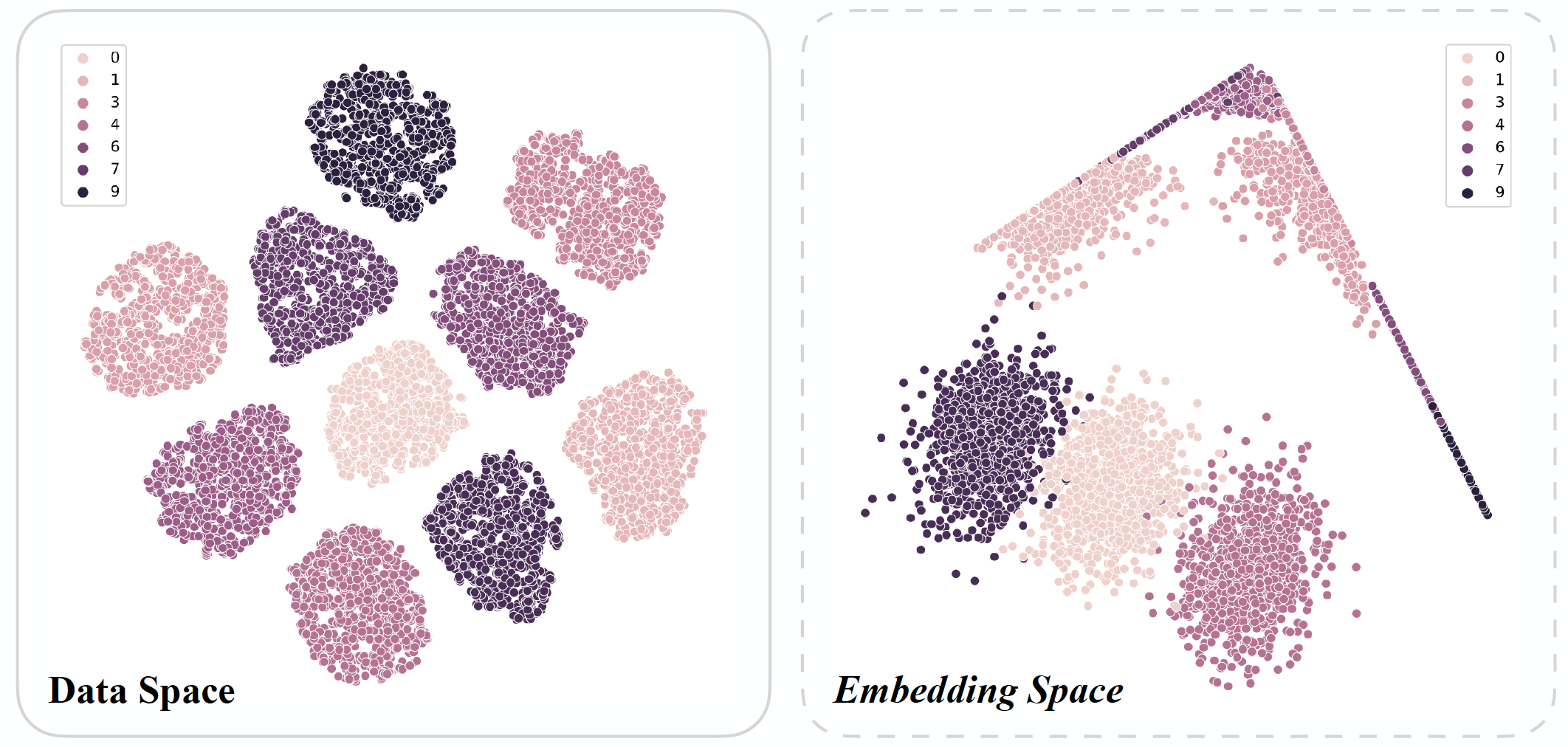}
\includegraphics[width=0.3\linewidth]{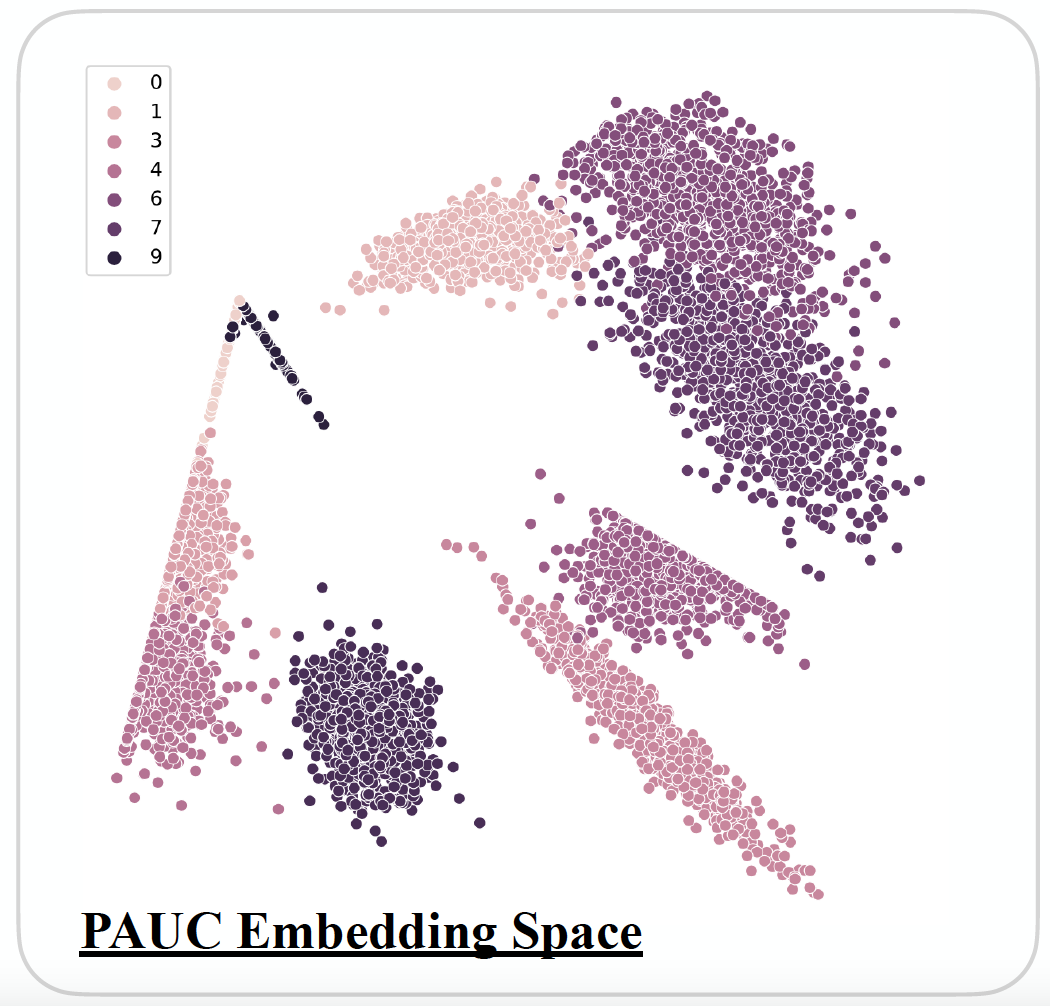}
\caption{Illustration of the ``coagulation'' phenomenon of prototypes in 2D-toy examples: (Top Left) Original data space; (Top Right) unnormalized CSL embedding space; (Bottom) unnormalized PAUC embedding space. The prototypes in the proposed PAUC embedding space are more proportioned then the CSL one. }
\vspace{-0.5cm}
\label{fig: title_img}
\end{figure}

In contrast to the instance-wise contrastive learning, where the InfoNCE loss~\cite{sohn2016improved} is computed between instance-level features, a typical prototypical contrastive self-supervised learning (CSL) framework calculates the contrastive loss between features at the cluster level. 
However, optimizing the prototype-level contrastive loss tends to lead to trivial solutions, termed \emph{prototype collapse}.
Instead, it would sacrifice the variety and diversity of features in the embedding space for the samples with similar semantic information.
Meanwhile, once a sample is assigned to an unrelated prototype during the optimization progress, the quality of learned representations further deteriorates since features are less informative except its prototype. 

To estimate the severity of the collapse, we treat the prototypes as independent probability distributions which describe the uncertainty of data being a certain prototype, and evaluate the quality of pre-trained representations at \emph{cluster-level}, on the \textbf{N}ormalized hyper-sphere by \textbf{E}arth \textbf{M}over’s \textbf{D}istance (NEMD).
Intuitively, prototypes containing different semantic information should uniformly distribute on a normalized hyper-sphere, with moderate overlapping (due to the ambiguity of boundary samples).
If they are collapsing, then the distance between them should increase, causing ``void'' on the hyper-sphere. 
This fact will be empirically analyzed in this work by experiments on the 2D embedding space, under the PCL~\cite{li2021prototypical} framework.
Subsequently, we tackle this collapsing issue by revising the vanilla PCL loss with three additional terms, dubbed,
\textbf{P}rototypical representation through \textbf{A}lignment, \textbf{U}niformity and \textbf{C}orrelation \textbf{(\texttt{PAUC})}.
Specifically,  we impose the prototypical alignment loss between positive prototypes to pull embeddings from positive prototypes together;
to achieve a uniform distribution of the prototypes on the normalized hyper-sphere, we then consider the prototypical uniformity loss between all prototypes; to increase the diversity and discriminability of prototypical level features, we further take the prototypical correlation loss motivated by mutual information into account.

We conduct extensive experiments and ablation studies on ImageNet-100 and ImageNet-1K datasets, and compare with state-of-the-art instance-wise and prototypical frameworks. The contributions in this work can be summarized as follows:
\vspace{-0.5em}
\begin{itemize}
    \item We demonstrate the \emph{prototype} collapse issue of pre-trained features in a typical PCL framework and analyze it with the proposed NEMD score defined on the normalized embedding space.
    \item We improve the prototypical learning of representations through alignment, uniformity, and correlation, mitigating the prototype collapse issue. 
    \item Extensive experiments on various benchmarks demonstrate the effectiveness of our method on enriching the diversity and alleviating the collapsing issue in meaningful prototypical representations.
\end{itemize}
\section{Related Work}
\vspace{-0.5em}
\noindent\textbf{Taxonomy of Self-supervised Learning.}
Self-supervised learning~\cite{wu2018unsupervised,tian2020what,chen2020simple,tian2021understanding} has achieved remarkable progress in recent years, in representation learning without the need for class label supervision.
The core part of self-supervised algorithms~\cite{zhao2021what,feichtenhofer2021a,purushwalkam2020demystifying} is to build a specific task for networks to learn. Typically, most algorithms utilize data augmentation to generate different views of an anchor sample. Then the networks are optimized to maximize the mutual information between different views.
The maximization can be achieved in both contrastive~\cite{chen2020simple,he2019moco} and non-contrastive approach~\cite{chen2021simsiam,zbontar2021barlow}. Likewise, the mutual information in contrastive framework can be also estimated in both instance-wise approach~\cite{chen2020simple,he2019moco,grill2020bootstrap} and prototypical approach~\cite{caron2020unsupervised,li2021prototypical,mo2021spcl}. Below, we provide related backgrounds on these two approaches.

\noindent\textbf{Instance-wise Contrastive Learning~(ICL).}
The goal of ICL~\mbox{\cite{chen2020simple,chen2020big,grill2020bootstrap,he2019moco,chen2020mocov2,chen2021simsiam}} is to bring the embedding of different views from the same instance closer, and push embeddings of views from different instances far apart using instance-level contrastive loss.
This is commonly achieved by a large batch size that can accumulate positive and negative pairs in the same batch~\cite{chen2020simple,chen2020big}, or a momentum encoder to update negative instances from a large and consistent dictionary on the fly~\cite{he2019moco,chen2020mocov2}. Other related works~\cite{wang2020hypersphere,chen2020intriguing} generalize the instance-wise contrastive loss to the alignment of representations from positive pairs and uniformity of the induced distribution of the normalized embeddings on the hyper-sphere.
We are motivated by these works yet explore those properties under the prototypical contrastive learning framework.

\noindent\textbf{Prototypical Contrastive Learning.}
Compared to the extensive ICL frameworks, there are fewer works focusing on the prototypical contrastive learning task. SwAV~\cite{caron2020unsupervised} proposes to simply predict the code of a view from the representation of the augmented view, where the code is obtained by multiplying the cluster assignments of the data with the same linear transformation matrix. PCL~\cite{li2021prototypical} replaces the InfoNCE loss with ProtoNCE loss for contrastive learning to encourage representations closer to their assigned prototypes and far from negative prototypes. CLD~\cite{wang2021cld} introduces the cross-level discrimination between instances and local instance groups to increase the positive/negative instance ratio of contrastive learning for better invariant mapping. SPCL~\cite{mo2021spcl} employs an offline prototype spawn approach with $k$-means clustering and regularizes samples to their corresponding prototypes explicitly. Although the prototypical framework demonstrates its superiority in classification downstream tasks, \textit{e.g.,} the ImageNet-1K, the inferior quality of features caused by the prototype collapse has been barely studied.

\begin{table}[!htb]
\setlength{\abovecaptionskip}{-0em}
\setlength{\belowcaptionskip}{-0em}
    \centering
	\renewcommand\tabcolsep{6.0pt}
	\caption{Comparison with existing works.}
    \label{tab: cmp_work}
	\scalebox{0.8}{
		\begin{tabular}{lccc}
			\toprule
			Method & Publication & Loss & Focus \\
			\midrule
			DIM~\cite{bachman2019amdim}  & NeurIPS'2019 & mutual information & instance \\
			MoCo+A/U~\cite{wang2020hypersphere} & ICML'2020 & align/uniform & instance \\
			VICReg~\cite{bardes2021vicreg} & ICLR'2022 & covariance & instance \\
			\textbf{\texttt{PAUC}} (ours) & -- & align/uniform/correlation & prototype\\
			\bottomrule
			\end{tabular}}
\end{table}

\noindent\textbf{Collapse in Contrastive Learning.}
A recent work~\cite{hua2021feature} discovers the feature collapse problem in self-supervised settings. The authors compartmentalize the collapse into two categories: \emph{complete collapse} when the network produces constant; and \emph{dimensional collapse} when some dimensions of the embedded feature representations are non-informative. Decorrelated batch normalization is then utilized to prevent the later one. 
Yet another concurrent work~\cite{bardes2021vicreg} in Table~\ref{tab: cmp_work} also aims to prevent collapsed solutions by explicitly applying regularization termed Invariance and Covariance. However, in this study, we focus on the \emph{prototype collapse}, a different type of collapse happens when the samples inside a prototype lose their diversity.

\vspace{-1em}
\section{Preliminary}
\vspace{-0.5em}
Under the self-supervised setting, there are two main branches of contrastive learning frameworks, including instance-wise contrastive learning (ICL)~\cite{chen2020simple,chen2020big,he2019moco,chen2020mocov2,cao2020parametric} and prototypical contrastive learning (PCL)~\cite{li2021prototypical,caron2020unsupervised,wang2021cld}. 
The ICL framework is optimized using an ordinary NT-Xent~(the normalized temperature-scaled cross-entropy loss) as the contrastive loss, called InfoNCE, to maximize the similarity between positive samples and minimize the similarity between negative samples. 
The PCL idea mainly focuses on applying a similar InfoNCE loss in terms of the prototypical level to distinguish positive and negative prototypes, where prototypes are defined under the clustering mechanism.
To explain them in a unified manner, we define notations as follows.

\noindent\textbf{Notations }
Given a set of training examples $\mathcal{X}=\{\mathbf{x}_1,\mathbf{x}_2, \cdots, \mathbf{x}_n\}$, a mapping function $f(\cdot)$ is applied to generate representations $\mathcal{V}=\{\mathbf{v}_1,\mathbf{v}_2, \cdots, \mathbf{v}_n\}$, \textit{i.e.,} $\mathbf{v}_i = f(\mathbf{x}_i),\forall{}i$. 
In the ICL setting, we denote the number of negative instances as $r$.
In the PCL framework, we denote the times of clustering as $M$, and the number of prototypes as $k_m, m\in \{1, 2, \cdots, M\}$. Therefore, we have a set of different number of prototypes $K = \{k_1, k_2, \cdots, k_M\}$.
The prototypes of the samples using $k_m$ clusters are marked as $\mathcal{C}^m=\{\mathbf{c}_1,\mathbf{c}_2, \cdots, \mathbf{c}_{k_m}\}$.

The InfoNCE~\cite{sohn2016improved,wu2018unsupervised,oord2018representation} objective of the instance-wise contrastive learning is defined by

\begin{equation}
    \mathcal{L}_{\text{InfoNCE}} = \sum_{i=1}^n -\log \frac{\exp(\mathbf{v}_i\cdot \mathbf{v}_i^\prime /\tau)}{\sum_{j=1}^{r}\exp(\mathbf{v}_i\cdot\mathbf{v}_j^\prime /\tau)}
\end{equation}
where $\mathbf{v}_i, \mathbf{v}_i^\prime, \mathbf{v}_j$ represent the anchor, positive, and negative embedding, respectively for each training sample $i$, and $\tau$ is a temperature hyper-parameter.
The operation $\cdot$ denotes the trivial inner product of two vectors.

In prototypical contrastive learning, we replace $r$ negative instances with $r$ negative prototypes to calculate the normalization term. 
For robust estimation, we average the prototypical probability by sampling $M$ steps with a set of different numbers of clusters $K$.
Thus, the prototypical InfoNCE objective with $\mathcal{L}_{\text{InfoNCE}}$ is defined as: 
\begin{equation}\label{eq:ProtoNCE}
\begin{aligned}
    \mathcal{L}_{\text{ProtoNCE}} & = \mathcal{L}_{\text{InfoNCE}} + \sum_{i=1}^n -\frac{1}{M}\sum_{m=1}^{M}\log \frac{\exp(\mathbf{v}_i\cdot \mathbf{c}_s^m /\phi_s^m)}{\sum_{j=1}^{r}\exp(\mathbf{v}_i\cdot \mathbf{c}_j^m /\phi_j^m)}
\end{aligned}
\end{equation}

where $\mathbf{v}_i$ is the anchor embedding for each training sample $i$, 
and $\mathbf{c}_s^m, \mathbf{c}_j^m$ are the positive prototype $s$ that the sample $i$ belongs to and the negative prototype $j$ at $m$ step, 
and $\phi_s^m, \phi_j^m$ are concentration estimation indicators for the distribution of embeddings around the prototype $s, j$ at $m$ step.
This term is used for pulling similar embeddings together around the target prototype.

\noindent\textbf{Normalized Earth Moving Distance~(NEMD) }
We present the \textit{Normalized Earth Mover’s Distance} to quantify the level of prototype collapse issue existing in current PCL methods. 
We start with the formulation of distance among prototypes on ``earth'', to calculate how far two prototypes are separated.
To achieve this, we assume the prototypes as empirical distributions supported on a normalized hyper-sphere. 
NEMD is then calculated 
between distributions associated with prototypes to measure the distance of separate prototypes generated in PCL, which is formally defined by
\begin{equation}
    \mathcal{L}_{\text{NEMD}}
    (p, q) = \inf_{\gamma \in\prod(\mathbf{c}_p^m, \mathbf{c}_q^m)}\mathbb{E}_{(x, y)\sim{\gamma}} \lVert x-y\rVert
\end{equation}
where $\prod(\mathbf{c}_p^m, \mathbf{c}_q^m)$ denotes the set of all joint distributions $\gamma(x,y)$ whose marginals are $\mathbf{c}_p^m$ and $\mathbf{c}_q^m$ associated with prototypes $p$ and $q$, respectively.
Here $\gamma(x,y)$ intuitively indicates how many ``clusters of samples'' should be transported from the embedding $x$ to the embedding $y$, in order to transform the prototype $\mathbf{c}_p^m$ into the prototype $\mathbf{c}_q^m$.
In the experiments, we adopt the Sinkhorn~\cite{marco2013sinkhorn} algorithm to calculate this score for fast computation of optimal transport between prototypes.
It is also worth mentioning that, in theory there exists the case when all the embedded features collapse to one point regardless of the prototypes, and NEMD thus equals 0 in this case.
However, in practice we barely fall into such trivial solutions due to the regularization of the prototype loss. Therefore, in this study we assume that NEMD is computed on well-optimized networks.
To evaluate the prototype collapsing problem, we conduct extensive analytical experiments on the synthetic 3D isotropic Gaussian data, embedding them into a 2D space. 
Specifically, we generate isotropic Gaussian blobs with various classes (10, 20, 30, 40, 50), and each class has 1000 samples. Note that the original dimension of data is 3 in this tiny experiment.
On the synthetic data we then carry out the vanilla PCL framework and its variant called PAUC with alignment, uniformity, and correlation (proposed in the next section), respectively.
To further evaluate the quality of prototypical solutions, we calculate the NEMD score between pre-trained prototypical embeddings and also visualize pre-trained embeddings by projecting them into a unit circle feature space.

\begin{figure*}[t]
\centering
  \includegraphics[width=0.98\linewidth]{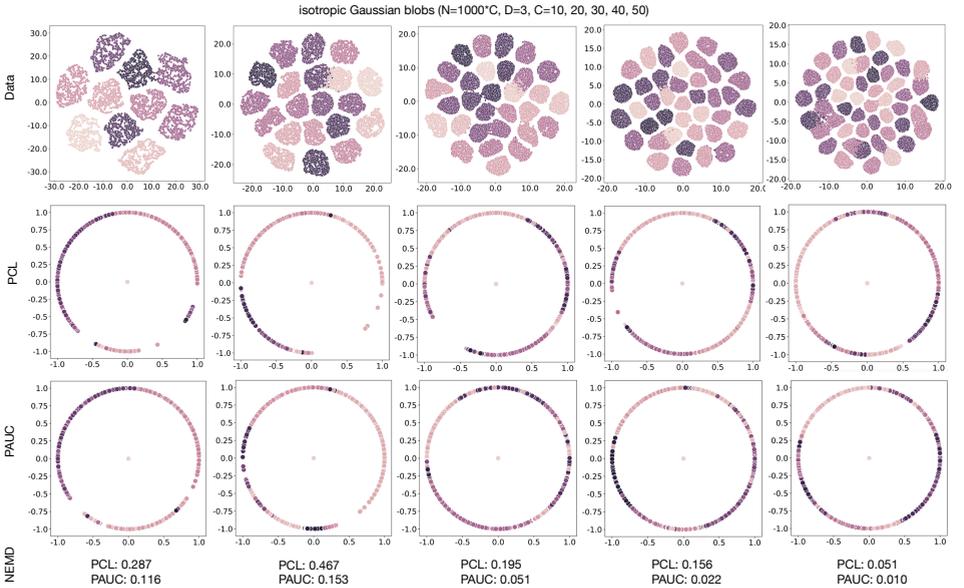}
  \vspace{-0.5em}
   \caption{Comparison results of tiny experiments between prototypical contrastive learning (PCL) and our PAUC in terms of \textbf{qualitative} (the projection of pre-trained embedding on a unit circle feature space) and \textbf{quantative} ($\mathcal{L}_{\text{NEMD}}$ score).}
\vspace{-0.5em}
\label{fig: exp_tiny}
\end{figure*}

As shown in Fig.~\ref{fig: exp_tiny}, we can observe that there exist collapsing solutions in representations generated by PCL (in the second row), leading to a high NEMD score. 
Our PAUC achieves a lower NEMD score compared to the general PCL framework under all class settings.
This implies that the pre-trained representations generated by PAUC are more uniformly distributed on the normalized hyper-sphere. 
Results in Fig.~\ref{fig: exp_tiny} also validate the uniformity of our prototypical representations on the normalized hyper-sphere. 

\vspace{-1em}
\section{\textbf{\texttt{PAUC}}: Prototypical Alignment, Uniformity and Correlation}\label{sec:method}
In this section, we elaborate on the three proposed properties of prototypes, \textit{i.e.,} alignment, uniformity, and correlation, for learning prototypical representations, based on the vanilla PCL framework. 

\noindent\textbf{Alignment} To pull embeddings from positive prototypes together and push negative prototypes away, the prototypical alignment loss is defined with the expected distance between positive prototypes:
\begin{equation}\label{eq:align}
    \mathcal{L}_{\text{p-align}} (\mathbf{c}^m; s)  = \mathbb{E}_{(p, q)\sim p_{\text{pos}}} [\lVert \mathbf{c}_p^m - \mathbf{c}_q^m \rVert_2^{s}]
\end{equation}
where $\mathbf{c}_p^m$, $\mathbf{c}_q^m$ are the \emph{positive} prototypes $p, q$ that the sample $i$ belongs to at $m$ step, $p_{\text{pos}}$ denotes the distribution of positive prototypes in the hyper-sphere, and $s$ is a positive factor to define the distance metric between $\mathbf{c}_p^m$ and $\mathbf{c}_q^m$. 

It is worth mentioning that, in the ordinary protoNCE loss as Eq.~\eqref{eq:ProtoNCE}, the alignment is achieved by matching other samples to the anchor. From the perspective of numerical optimization, when reducing over different number of prototypes, all the samples from positive prototypes are pulled together to their anchor sample, which might result in unwanted ``coagulation'' effect in a long-term training progress.

\noindent\textbf{Uniformity} To alleviate the inter-prototype collapsing issues, we adopt a uniformity loss to learn a uniform distribution of the prototypical features on the hyper-sphere.
Motivated by the uniformity property proposed in ICL, we consider the Gaussian potential kernel~\cite{bartok2015gaussian} and define the prototypical uniformity loss by
\begin{equation}\label{eq:uniform}
    \mathcal{L}_{\text{p-uniform}} (\mathbf{c}^m; t)  = \mathbb{E}_{(p, q)\sim p_{\text{proto}}} [e^{-t\lVert\mathbf{c}_p^m - \mathbf{c}_q^m\rVert_2^{2}}]
\end{equation}
where $\mathbf{c}_p^m$, $\mathbf{c}_q^m$ are the prototypes $p, q$ that the sample $i$ belongs to at $m$ step, $p_{\text{proto}}$ denotes the distribution of all prototypes in the hyper-sphere, and $t$ is a positive factor to define the weight of the $\ell_2$ distance between $\mathbf{c}_p^m$ and $\mathbf{c}_q^m$.

\noindent\textbf{Correlation} In order to distinguish the difference between each prototype further to avoid inter-prototype collapsing,
we borrow the idea of mutual information and define the correlation loss between positive prototypes and negative prototypes by
\begin{equation}\label{eq:corr}
    \mathcal{L}_{\text{p-corr}} = 
    \mathbb{E}_{(p, q)\sim p_{\text{proto}}} [\mathbf{c}_p^m\log \left(\mathbf{c}_q^m \odot (\mathbf{c}_p^m)^{-1}\right)]
\end{equation}
where $\mathbf{c}_p^m$, $\mathbf{c}_q^m$ are prototypes $p, q$ that the sample $i$ belongs to at $m$ step, and $\odot$ is for element-wise product. Intuitively, this loss term drives the samples from different prototypes being unrelated among all the dimensions of the embedding space.

\noindent\textbf{Overall Loss} 
The overall loss $\mathcal{L}$ is therefore defined with the weighted summation of the three losses by
\begin{equation}\label{total_loss}
\begin{aligned}
    \mathcal{L} &= \mathcal{L}_{\text{InfoNCE}}  - \sum_{i=1}^n \frac{1}{M}\sum_{m=1}^{M}\Big(\log \frac{\exp(\mathbf{v}_i\cdot \mathbf{c}_s^m /\phi_s^m)}{\sum_{j=1}^{r}\exp(\mathbf{v}_i\cdot \mathbf{c}_j^m /\phi_j^m)} 
    + \alpha \mathcal{L}_{\text{p-align}} + \beta \mathcal{L}_{\text{p-uniform}} +
    \gamma \mathcal{L}_{\text{p-corr}}\Big)
\end{aligned}
\end{equation}
where $\alpha, \beta$, and $\gamma$ denote the weights of the alignment, uniformity and correlation loss, respectively.
We conduct extensive ablation studies in the supplementary material to explore the effects of each loss on the quality of prototypical representations generated by our PAUC.

\vspace{-1em}

\section{Experiments}\label{sec:exp}
\vspace{-0.5em}
\subsection{Dataset and Configurations}\label{sec:dataset}
\vspace{-0.5em}
Following previous methods~\cite{li2021prototypical,caron2020unsupervised,chen2020simple,tian2020contrastive,he2019moco}, we employ two benchmark datasets (ImageNet-100 and ImageNet-1K) for evaluating classification performance.
For a fair comparison with state-of-the-arts, we adopt a ResNet-50~\cite{he2016resnet} as the encoder, where the last fully connected layer outputs a 128-D embedding with $\ell_2$-normalization.
Same as PCL~\cite{li2021prototypical}, we apply data augmentation methods with the random crop, random color jittering,
random horizontal flip, and random grayscale conversion.

For ImageNet-100 pre-training, we set number of clusters $K = {2500, 5000, 10000}$, $r=1024$.
We apply SGD as our optimizer, with a weight decay of 0.0001, a momentum of 0.9, and a batch size of 256.
We train for 200 epochs and use the first 20 epochs as a warm-up step using only the InfoNCE loss.
We set the initial learning
rate as 0.03, and multiply it by 0.1 at 120 and 160 epochs.

For ImageNet-1K,
we set $K = {25000, 50000, 100000}$, $r=16000$. 
For other hyper-parameters, we follow the same setting as ImageNet-100 pre-training.
For an efficient $k$-means clustering, we adopt the faiss library~\cite{JDH17faiss} during the pre-training.
The whole training time for ImageNet-1K is 132 hours using 8 Tesla V100 GPUs, and 15 hours for ImageNet-100.

\begin{table*}[t]
	\renewcommand\tabcolsep{6.0pt}
	\centering
	\caption{Linear classification on ImageNet-100. Bold and underline denote the first and second place.}
	\label{tab: cls_imagenet100}
	\scalebox{0.75}{
		\begin{tabular}{lllllll}
			\toprule
			\textbf{Method} & \textbf{Arch.} & \textbf{Param.(M)} & \textbf{Batch} & \textbf{Epochs} & \textbf{Top-1(\%)} & \textbf{Top-5(\%)} \\
			\midrule
        	CMC\cite{tian2020contrastive}  & ResNet-50 & $24$ & $256$ & $200$ & $66.20$ & $87.00$ \\
			MoCo\cite{he2019moco} & ResNet-50 & $24$ & $256$ & $200$ & $72.80$ & $91.64$ \\
			Biased CMC \cite{chuang2020debiased} & ResNet-50 & $24$ & $256$ & $200$ &  $73.58$ & $92.06$ \\
			Debiased CMC \cite{chuang2020debiased} & ResNet-50 & $24$ & $256$ & $200$ &  $74.60$ & $92.08$ \\
        	MoCo+align/uniform\cite{wang2020hypersphere} & ResNet-50 & $24$ & $256$ & $200$ &  $74.60$ & $92.74$ \\
        	NPID~\cite{wu2018unsupervised} & ResNet-50 & $24$ &$256$ & $200$ & $75.30$ & $-$ \\
        	BYOL~\cite{grill2020bootstrap} & ResNet-50 & $24$ & $4096$ & $1000$ & $75.80$ & $-$ \\
			PCL-v1~\cite{li2021prototypical} & ResNet-50 &$24$ & $256$ & $200$ & $76.17$ & $93.52$ \\
			PCL-v2~\cite{li2021prototypical} & ResNet-50+MLP &$28$ & $256$ & $200$ & $78.35$ & $94.25$ \\
			SwAV~\cite{caron2020unsupervised} & ResNet-50 & $24$ & $4096$ & $200$ &$80.20$ & $95.02$ \\
        	LooC~\cite{xiao2021what} & ResNet-50 & $24$ & $256$& $200$ &$81.10$ & $95.30$ \\
			CLD~\cite{wang2021cld} & ResNet-50 & $24$ & $256$ & $200$& \underline{$81.50$}& \underline{$95.48$} \\
			\textbf{\texttt{PAUC}} (ours) & ResNet-50 & $24$ & $256$ & $200$ & $\bm{84.46}$ & $\bm{97.15}$ \\
			\bottomrule
			\end{tabular}}
\end{table*}
\vspace{-0.5em}

\begin{figure*}[!htb]
\centering
  \includegraphics[width=\linewidth]{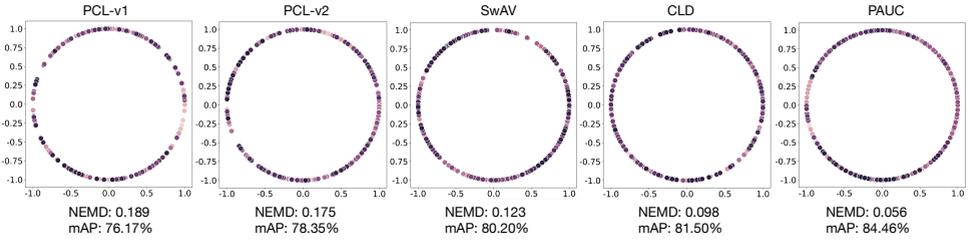}
  \vspace{-1.5em}
   \caption{Comparison results on ImageNet-100 dataset between current prototypical contrastive learning methods (PCL-v1, PCL-v2, SwAV, CLD) and our PAUC in terms of \textbf{qualitative} ( pre-trained embeddings on a unit circle feature space) and \textbf{quantative} ($\mathcal{L}_{\text{NEMD}}$ and top-1 accuracy).}
\label{fig: exp_vis_100}
\end{figure*}

\vspace{-0.5em}
\subsection{Comparison with State-of-the-arts}\label{sec:sota}
\vspace{-0.5em}
\noindent\textbf{ImageNet-100.}
We evaluate the linear classification for the ImageNet-100 dataset, where linear models are trained on frozen features from different self-supervised methods shown in Table~\ref{tab: cls_imagenet100}. Our PAUC substantially outperforms existing methods in terms of both instance-wise and prototypical contrastive learning. 
We achieve new state-of-the-art performance for linear classification on the ImageNet-100 dataset.  
This demonstrates the effectiveness of our PAUC pre-trained representations in transfer learning on image classification.

\begin{table*}[!htb]
	\renewcommand\tabcolsep{3.0pt}
	\centering
	\caption{Top-1 accuracy for linear classification on ImageNet-1K, where models are trained on frozen features from different methods. Bold and underline numbers denote the first and second place.} 
	\label{tab: cls_imagenet}
	\scalebox{0.8}{
		\begin{tabular}{llcccc}
			\toprule
			\textbf{Method} & \textbf{Arch.} & \textbf{Param.(M)} & \textbf{Batch} & \textbf{Epochs} & \textbf{Top-1(\%)} \\
			\midrule
			\multicolumn{5}{l}{\textit{Instance-wise Contrastive:}} \\
			CPC~\cite{oord2018representation} & ResNet-101 & $28$ & $512$ & $200$ & $48.70$ \\
			MoCo~\cite{he2019moco} & ResNet-50 & $24$ & $256$ & $200$ & $60.60$ \\
			PIRL~\cite{misra2020self} & ResNet-50 & $24$ & $1024$ & $800$  & $63.60$ \\
			CMC~\cite{tian2020contrastive} & ResNet-50+MLP$\{L, ab\}$ & $47$ & $256$ &  $200$ & $64.00$ \\
			CPCv2~\cite{henaff2019cpcv2} & ResNet-170 & $303$ & $512$ & $200$ & $65.90$ \\
			MoCo+align/uniform~\cite{grill2020bootstrap} & ResNet-50 & $24$ & $256$ &  $200$ & $67.69$ \\
			AMDIM~\cite{bachman2019amdim} & Custom-ResNet & $192$ & $1008$ & $150$ & $68.10$ \\
		    LoCo~\cite{xiong2020loco} & ResNet-50 & $24$ & $4096$ & $800$ & $69.50$ \\
		    SimCLR~\cite{chen2020simple}  & ResNet-50+MLP & $28$ & $4096$ & $400$ & $70.00$ \\
    		InfoMin~\cite{tian2020what} & ResNet-50 & $24$ & $256$ & $200$ & $70.10$ \\
    		MoCHi~\cite{kalantidis2020hard} & ResNet-50+MLP & $28$ & $512$ & $200$ & $70.60$ \\
		    PIC~\cite{cao2020parametric} & ResNet-50 & $24$ & $512$ &$1600$ & $70.80$ \\
		   SWD~\cite{chen2020intriguing} & ResNet-50+MLP & $28$ & $2048$ &$800$ & $70.90$ \\
		   MoCov2~\cite{chen2020mocov2} & ResNet-50+MLP & $28$ & $256$ & $200$ & $71.10$ \\
			SimSiam~\cite{chen2021simsiam} & ResNet-50+MLP & $28$ & $256$ & $800$ & $71.30$ \\
			SimCLRv2~\cite{chen2020big} & ResNet-50+MLP & $28$ & $4096$ & $800$ & $71.70$ \\
			MoCov3~\cite{chen2021mocov3} & ResNet-50+MLP & $28$ & $4096$ &$300$ & $72.80$ \\
                VICReg~\cite{bardes2021vicreg} & ResNet-50+MLP & $28$ & $2048$ & 1000 & $73.20$ \\
			AdCo~\cite{hu2021adco} & ResNet-50 & $24$ & $256$ &$200$ & $73.20$ \\
		    Barlow Twins~\cite{zbontar2021barlow} & ResNet-50 & $24$ & $2048$ & $1000$ & $73.20$ \\
		    BYOL~\cite{grill2020bootstrap} & ResNet-50+MLP & $35$ & $4096$ & $400$ & $73.20$ \\
		    BYOL~\cite{grill2020bootstrap} & ResNet-50+MLP & $35$ & $4096$ & 1000 & $74.30$ \\
		    MoCov3~\cite{chen2021mocov3} & ResNet-50+MLP & $28$ & $4096$ &$1000$ & \underline{$74.60$} \\

		    \midrule
            \multicolumn{5}{l}{\textit{Prototypical Contrastive:}} \\
		    PCL~\cite{li2021prototypical} & ResNet-50 & $24$ & $256$ & $200$ & $61.50$\\
		    PCLv2~\cite{li2021prototypical} & ResNet-50+MLP & $28$ & $256$ & $200$ &  $67.60$ \\
		    CLD~\cite{wang2021cld} & ResNet-50 & $24$ & $256$ &$200$ & $71.50$ \\
		     SwAV~\cite{caron2020unsupervised} & ResNet-50+MLP & $28$ & $256$ & $200$& $72.70$ \\
	    	\textbf{\texttt{PAUC}} (ours) & ResNet-50 & $24$ & $256$& $200$ & $\bm{75.16}$  \\
			
			
			\bottomrule
			\end{tabular}}
\end{table*}

Additionally, we report the comparison results with current prototypical contrastive learning methods (PCL-v1~\cite{li2021prototypical}, PCL-v2~\cite{li2021prototypical}, SwAV~\cite{caron2020unsupervised}, CLD~\cite{wang2021cld}) in terms of qualitative and quantitative aspects.
Notably, we project pre-trained embeddings generated by different methods on a normalized unit circle feature space and compare the $\mathcal{L}_{\text{NEMD}}$ score and top-1 accuracy in Fig.~\ref{fig: exp_vis_100}. 
Our PAUC achieves the lowest $\mathcal{L}_{\text{NEMD}}$ score while performing the best linear classification on ImageNet-100. 
This validates the rationality of our proposed $\mathcal{L}_{\text{NEMD}}$ in evaluating the quality of prototypical representations. 
As can be seen in Fig.~\ref{fig: exp_vis_100}, our PAUC pre-trained representations achieve better uniformity on the normalized hyper-sphere compared to previous methods, which validates the effectiveness of our PAUC in mitigating collapsing issues for the general PCL framework.

\noindent\textbf{ImageNet-1K.}
Following previous self-supervised methods~\cite{li2021prototypical,caron2020unsupervised,wang2021cld,chen2020simple,tian2020contrastive,he2019moco}, we compare the top-1 accuracy for linear classification on ImageNet-1K dataset as shown in Table~\ref{tab: cls_imagenet}.
Our PAUC achieves the best results compared to existing prototypical contrastive learning methods. 
Under the same setting, our PAUC outperforms SwAV~\cite{caron2020unsupervised} by a large margin, \textit{i.e.}, by $2.46\%$.
This validates the advantage of our PAUC pre-trained representations on transfer learning for image classification again.
Furthermore, it also demonstrates that, 
addressing the collapse is beneficial for improving the quality of pre-trained representations for downstream tasks. 
We also evaluate our PAUC pre-trained representations on object detection and report the detailed comparison in the supplementary material. 
Our PAUC achieves better or comparable performance compared to existing approaches.

When compared to ICL learning methods, PAUC pre-trained with small batch size and epochs also outperforms BYOL~\cite{grill2020bootstrap} with large batch size and pre-training epochs. 
This further implies that our PAUC has great potential in saving much pre-training time and GPU memories for contrastive self-supervised learning.
This also validates the advantage of prototypical contrastive learning frameworks over the instance-wise contrastive learning methods.

\begin{figure*}[t]
\setlength{\abovecaptionskip}{-1em}
\setlength{\belowcaptionskip}{-0.5em}
\centering
  \includegraphics[width=\linewidth]{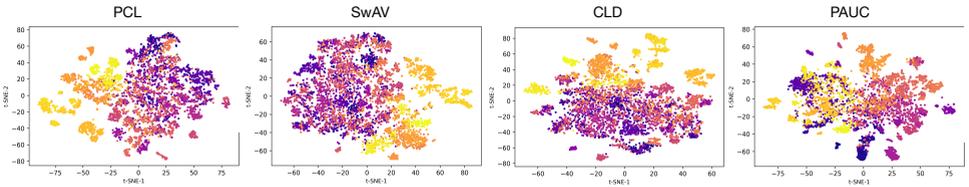}
  \vspace{-1em}
   \caption{Visualization of pre-trained representations generated by PCL, SwAV, CLD, and PAUC (ours) from random $100$ classes in the ImageNet validation set.}
\label{fig: exp_vis}
\vspace{-0.5em}
\end{figure*}

\noindent\textbf{Visualization of Pre-trained Representations.}
To better evaluate the quality of pre-trained representations, we visualize the self-supervised learned embeddings generated by different approaches (PCL~\cite{li2021prototypical}, SwAV~\cite{caron2020unsupervised}, CLD~\cite{wang2021cld}) in Fig.~\ref{fig: exp_vis}.
Typically, we project the $2048$-dimensional embeddings pre-trained from ResNet-50 onto $2$ dimension using tSNE~\cite{laurens2008tsne} tools, and we randomly select  $100$ classes in ImageNet-1K validation set for visualization. 
Compared to the pre-trained representations generated by previous PCL frameworks, our PAUC pre-trained representations form more separated clusters, which are distributed more uniformly on the space in terms of all classes.
This further validates the effectiveness of our PAUC in improving the quality of prototypical representations.

\vspace{-0.5em}
\section{Conclusion}
\vspace{-0.5em}
In this work, we reveal the prototype collapse problems existing in the general prototypical contrastive learning settings.
We propose a Normalized Earth Mover’s Distance~(NEMD) between prototypes to measure the quality of pre-trained representations at the prototypical level.
To improve the quality of prototypical representations further, our model is trained to learn the alignment of prototypes, the uniformity of prototypes on the normalized hyper-sphere, and the correlation between prototypes.
Extensive experiments on tiny experiments and state-of-the-art benchmarks validate the effectiveness of our method on pre-training better prototypical representations.

\section*{Acknowledgement}
This work was partially supported by JSPS KAKENHI (Grant No. 20H04249, 20H04208) and the National Natural Science Foundation of China (Grant No. 62006045).

\bibliography{reference}
\end{document}